\documentclass{article}
\usepackage{subfig}

\usepackage{microtype}
\usepackage{graphicx}
\usepackage{booktabs} % for professional tables
\usepackage{multirow} 

\usepackage{hyperref}

\usepackage[accepted]{mlsys2024}

\mlsystitlerunning{In-Depth Analysis on Caching and Pre-fetching in Mixture of Experts Offloading}

\begin{document}

\twocolumn[
\mlsystitle{In-Depth Analysis on Caching and Pre-fetching in Mixture of Experts Offloading}

% unless you've provided the [accepted] option to the mlsys2024


\mlsyssetsymbol{equal}{*}

\begin{mlsysauthorlist}
\mlsysauthor{Shuning Lin}{equal,to}
\mlsysauthor{Yifan He}{equal,to}
\mlsysauthor{Yitong Chen}{equal,to}
\end{mlsysauthorlist}

\mlsysaffiliation{to}{Carnegie Mellon University, Pittsburgh, PA, USA}
\mlsyscorrespondingauthor{Shuning Lin}{shuningl@alumni.cmu.edu}
\mlsyscorrespondingauthor{Yifan He}{yifanhe@andrew.cmu.edu}
\mlsyscorrespondingauthor{Yitong Chen}{yitongc@alumni.cmu.edu}

\mlsyskeywords{Machine Learning, MLSys}

\vskip 0.3in

% \begin{abstract}
% Do we need abstract??

% Still need one
% \end{abstract}
]



\printAffiliationsAndNotice{\mlsysEqualContribution} % otherwise use the standard text.

\section{Introduction}

% \section{Overview}
\label{submission}

In today's landscape, Mixture of Experts(MoE) \cite{shazeer2017outrageously} is a crucial architecture that has been used by many of the most advanced models. One of the major challenges of MoE models is that they usually require much more memory than their dense counterparts due to their unique architecture \cite{rajbhandari2022deepspeed}, and hence are harder to deploy in environments with limited GPU memory, such as edge devices. MoE offloading \cite{eliseev2023fastInference} is a promising technique proposed to overcome this challenge, especially if it is enhanced with caching and pre-fetching, but prior work stopped at suboptimal caching algorithm and offered limited insights. In this work, we study MoE offloading in depth and make the following contributions: 1. We analyze the expert activation and LRU caching behavior in detail and provide traces. 2. We propose LFU caching optimization based on our analysis and obtain strong improvements from LRU. 3. We implement and experiment speculative expert pre-fetching, providing detailed trace showing its huge potential . 4. In addition, our study extensively covers the behavior of the MoE architecture itself, offering information on the characteristic of the gating network and experts. This can inspire future work on the interpretation of MoE models and the development of pruning techniques for MoE architecture with minimal performance loss.

\section{Related Work}

\subsection{Mixture of Experts Background}

In today's landscape, Mixture of Experts(MoE) \cite{shazeer2017outrageously} is a crucial architecture that has been used by many of the most advanced models,
for example Gemini 2.5 Pro \cite{comanici2025gemini} and Qwen3-1\cite{yang2025qwen3} clearly claimed they are based on MoE. Although GPT-5\cite{gpt5Website} and Grok-4\cite{grok4Website} did not disclose their architecture details, they are widely believed to also use MoE.

Significant strides have been made towards optimizing computational efficiency and resource allocation for MoE models. The introduction of ZeRO-Offload\cite{ren2021zero} and the advancements in MoE inference and training in DeepSpeed-MoE \cite{rajbhandari2022deepspeed} represent critical efforts towards democratizing access to large-scale MoE models. These contributions are crucial and demonstrate the potential of offloading strategies in improving computational efficiency. However, they predominantly focus on training phases and large-scale deployments, leaving a gap in research related to inference phase optimization in severely resource-limited environments such as in Edge Computing. \cite{eliseev2023fastInference} dived into fast inference mechanisms for MoE language models, introducing expert offloading enhanced by LRU cache. We use this work as our baseline.

The evaluation framework for massive multitask language understanding presented by Hendrycks et al. (2020) provides a solid benchmark for assessing the effectiveness of language models across a broad spectrum of tasks. We use this industry standard benchmark as our inference workload to test inference speed with our optimized caching algorithm.

\subsection{MoE Offloading}

In a transformer-based Mixture-of-Experts (MoE) architecture, the feed-forward network (FFN) layer is replaced by an MoE layer. The MoE layer is composed of multiple FFNs known as \textit{expert}s and a special gating network that routes input tokens to only some of the experts. This unique architecture makes the total size of MoE models very large as all experts are included in the model, but a much smaller number of parameters are actually used during inference because only one or few experts are selected(activated) for each token \cite{lepikhin2020gshard}.

If one does not have enough GPU memory to hold the entire model, it is simply not possible to run the model directly. Fortunately, Offloading is a solution to this problem and it has been used in training with limited GPUs for a long time \cite{ren2021zero}. During forward pass, the model parameters can be loaded into GPU on-demand: they are stored in the main memory by default, and only sent to GPUs when they are needed for the forward pass. For MoE particularly, this method has more potential because we only need to load the activated experts to GPU, hence we have more space on GPU wihch can be leveraged for further optimizations.

Obviously, the challenge of MoE offloading is the frequent transfer of expert parameters, which can cause long latency due to bandwidth limit. Compared to the time spent on waiting for parameter loading, the time spent on actual computation may be much shorter. Therefore, to reduce the latency of inference when using offloading for MoE models, it is important to reduce the time spent on waiting for expert loading.

Given the leverage of GPU space and the bottleneck of transfer time, caching is an appealing choice since it enables utilizing the space to pre-store parameters that can be immediately used for inference hence eliminate the time waiting for loading from memory.

\section{Methodology}

\subsection{Temporal Locality and Caching}

The Mixtral 8x7B paper \cite{jiang2024mixtral} presents an important finding: tokens next to each other tend to select the same experts. The probability for a token to select the same expert as its previous token is higher than random sampling(12.5\% in the case of 8 experts), sometimes near 30\%. \cite{eliseev2023fastInference}  utilized the temporal locality observed to propose caching for expert offloading, which significantly reduced the time spent on waiting for expert loading. The key idea is that by caching previously-used experts, if the same expert is selected again for the next token(recall temporal locality), the cached experts can be directly used rather than being fetched from the main memory. The source code of this work is available at \url{https://github.com/dvmazur/mixtral-offloading}.  We use this work as our baseline, build on top of its codebase to run analysis, and implement new methods.

\begin{figure}[h]
\includegraphics[width=8cm]{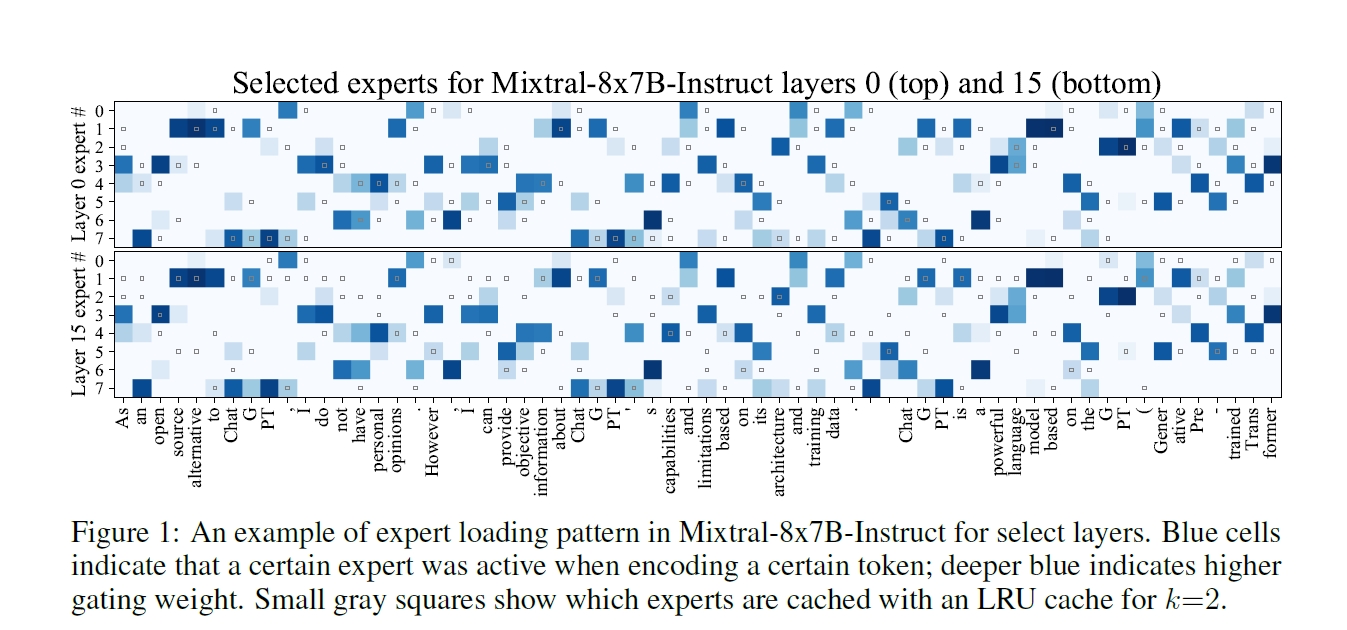}
\caption{The trace of LRU cache performance. The small gray squares(cached experts) tend to repeat history rather than predict the selected experts in the future.}
\centering
\label{fig1}
\end{figure}

The baseline work uses a simple LRU policy for cache eviction. It provides a single trace(Figure \ref{fig1}), where we can see there are a lot of cache misses (activated expert not cached) and ``miscached'' experts(cached expert not activated). However, it is very hard to draw conclusion or propose optimization from just a single trace. We investigate the implementation and build a tracing system, which can collect and visualize the entire activation and caching history at any layer, for any token, in any prompt. With this information, we are able to analyze the real performance of LRU caching. We dive deep into our traces and find there is an obvious expert imbalance phenomenon, which corresponds to what has been found in previous work\cite{lepikhin2020gshard}. To address this, we propose frequency-based LFU and obtain considerable improvements. Details are discussed in section 4.

\subsection{Speculative Expert Loading}
\cite{eliseev2023fastInference} also proposes a technique called \textit{speculative expert loading}: ``transformer layers are residual, i.e. each layer adds to the previous hidden states instead of re-computing them from scratch", therefore we can ``get an accurate guess of next layer’s experts by applying next layer’s gating function to previous layer’s hidden states". However, the original work did not provide the implementation of this method. We implement this method based on their description and measure the performance of this method, obtaining very positive results.

\section{Experiments}

\begin{table*}[t]
\centering
\resizebox{0.75\textwidth}{!}{%
\begin{tabular}{ccccc}
\toprule
\# Offloads per layer & MMLU (\%) & Tokens Per Second & Peak Memory (MB) \\ \midrule
% Hugging Face accelerate & - & - & - \\
4 & 63.16 & 4.23 & 11148.3 \\
5 & 61.40 & 4.78 & 9145.8 \\ 
6 & 59.65 & 7.16 & 7127.7 \\
\bottomrule
\end{tabular}%
}
\caption{The model performance by changing the number of offloads per layer using LRU Caching.}
\label{table-metrics}
\end{table*}

\subsection{Experiment Setup}

We first conducted experiments to evaluate the baseline cache and offloading techniques in terms of both inference speed and accuracy. To measure the speed of the MoE model, we set the batch size to 1 to do inference, and we used tokens generated per second as the metric. We also report the peak memory usage. To evaluate the accuracy, we leveraged the multiple-choice questions from the language comprehension and reasoning benchmark MMLU \cite{hendrycks2020measuring}, and performed 4-shot Chain-of-Thought prompting. The same CoT exemplars are taken from \cite{pitis2023boosted}, and we follow the answer cleansing procedures from Self-Consistency \cite{wang2022self}. Considering the large size of the full MMLU dataset, we used a subset of 57 samples, taking 1 sample from each of MMLU's 57 subjects. During autoregressive decoding, we followed the baseline to set both the temperature and top\_p to 0.9. Experimental results and analysis are in section \ref{lru-cache}.

During our analysis of the baseline results, we found that the distribution of activated experts is uneven, and the frequency-based LFU method for expert caching might bring a potential performance boost. We then implemented LFU caching and conducted experiments.We chose to do inference experiments on the prompt we define across a variety of hardware configurations, including NVIDIA A100, A6000, L40, and RTX 3090, in order to assess the scalability of our proposed method across a wide range of computational environments. To ensure the generated response content and length are comparable, we set both the temperature and top\_p to 0.1. Experimental results and analysis are in section \ref{lfu-cache}.

Additionally, we performed an in-depth analysis of caching activation history to gain insights into the efficiency of memory utilization and the effectiveness of caching strategies in enhancing model performance. We will discuss this in section \ref{analysis}.

\subsection{LFU}

LFU (Least Frequently Used) is a common caching algorithm used to manage a finite cache of resources. LFU caching algorithm evicts the least frequently accessed items first. In practice, we added one usage count field in the implementation of the information of experts.

The primary objective is to determine whether LFU caching can improve the computational efficiency and response time of a MoE system by preferentially retaining frequently accessed expert models in the cache.

We evaluated both LRU and LFU caching algorithm by comparing which experts they choose to store and which experts are truly selected to compute precision and recall, as well as the inference speed.

\subsection{Speculative Preloading}

We implemented the algorithm of speculative expert preloading by multiplying the hidden states at the current layer with the gating network at the next layer. When the model is initialized, each layer stores not only its own gating network, but also next layer's gating network. During inference, the hidden states obtained after the multi-head attention block is multiplied with the next layer's gating networks. Gating network is essentially a linear layer of shape [len\_hidden, num\_experts], therefore by multiplying hidden states with gating network then applying a softmax followed by a topK we can get K selected experts(in our case, K=2). We logged the two experts selected for next layer as well as the truly activated experts when forward computation arrives the next layer(where we can use next layer's true hidden states to get the real expert selection). 

We compared the full history of truly activated experts and speculatively guessed experts, computed the precision and recall. Note that we only implemented the algorithm of speculative loading hence and did not enable it to load experts due to complexity discussed in \ref{takeways}, hence we did not evaluate inference speed for this optimization.

\section{Analysis}
\label{analysis}

\subsection{LRU Cache Performance}
\label{lru-cache}

For the experiment in Table \ref{table-metrics}, we used the Mistral-8x7B-Instruct model on an NVIDIA A6000 GPU. We applied 4-bit HQQ quantization for the shared attention layers with a group size of 64, and 2-bit HQQ quantization for experts with a group size of 16. Compress zero is applied to all layers. We tested the model performance of by varying the number of offloads per MoE layer. We observed that the peak memory usage is of a linear relationship with the number of offloads, with about 2000 MB decrease every more offload. While increasing the offload numbers boosts the speed of token generation and decreases the required memory to perform the operations, the performance drops.

We recorded the full caching activation history. Here, we report the activation history obtained for the conversation below: 

\textit{Prompt: ``Introduce yourself, limit your response in 50 words.''}

\textit{Response: ``I am Pabla, a helpful and polite assistant, always eager to provide accurate and concise information in a timely manner. I'm an expert in finding quick answers to various types of questions, ensuring reliability and relevance in my responses. I am always available to assist you.''}

\begin{figure}[h]
\includegraphics[width=8cm]{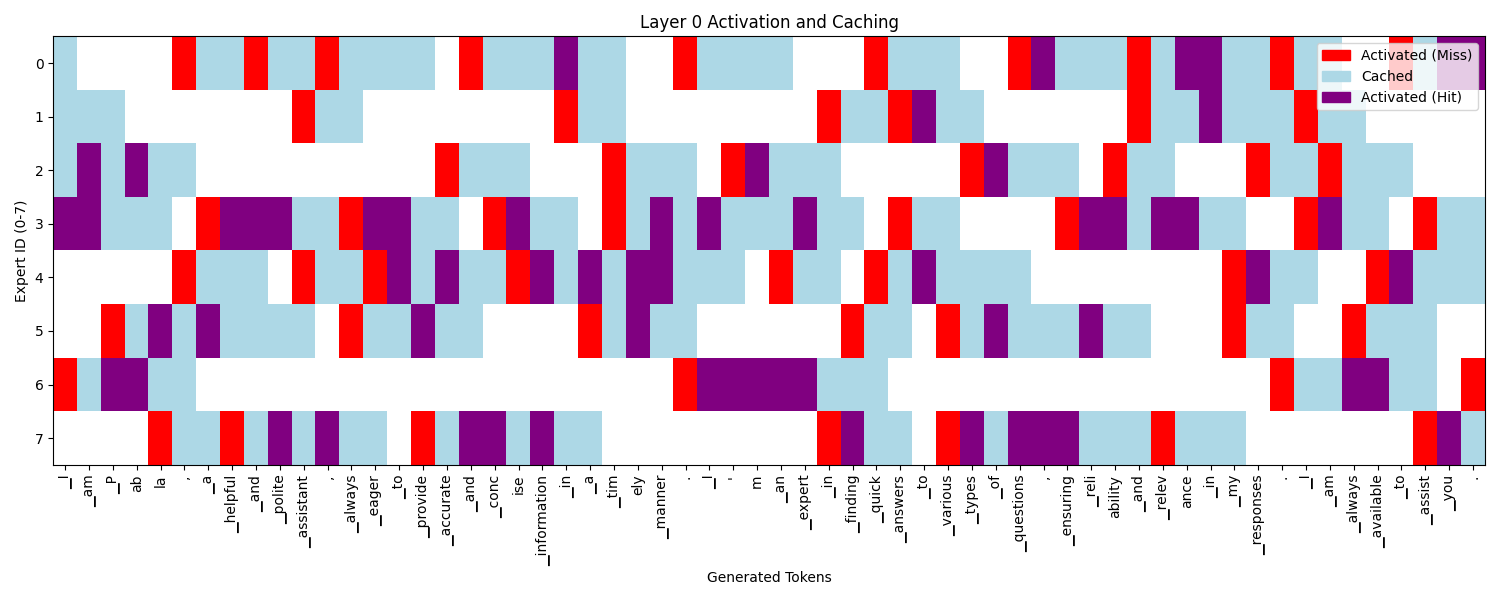}
\caption{The trace of expert activation and LRU cache with cache size=4 for the 1st layer. }
\centering
\label{p1l0}
\end{figure}

\begin{figure}[h]
\includegraphics[width=8cm]{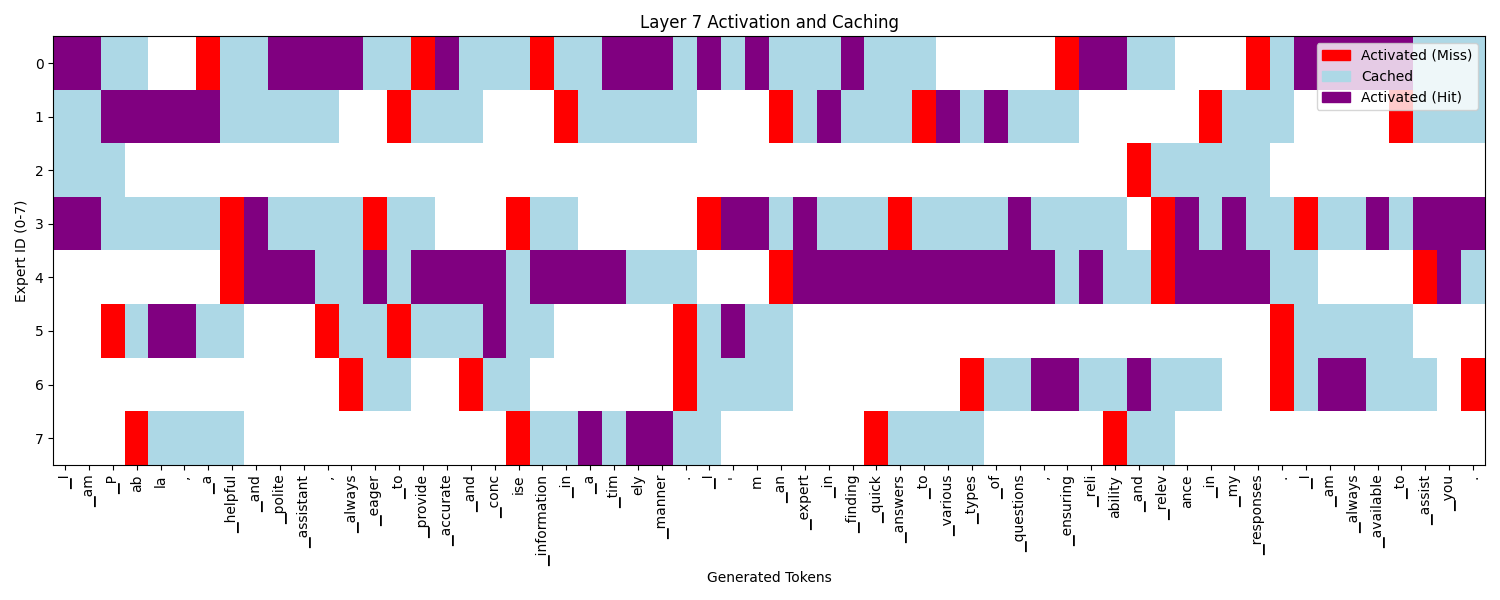}
\caption{The trace of expert activation and LRU cache with cache size=4 for the 8th layer. }
\centering
\label{p1l7}
\end{figure}

\begin{figure}[h]
\includegraphics[width=8cm]{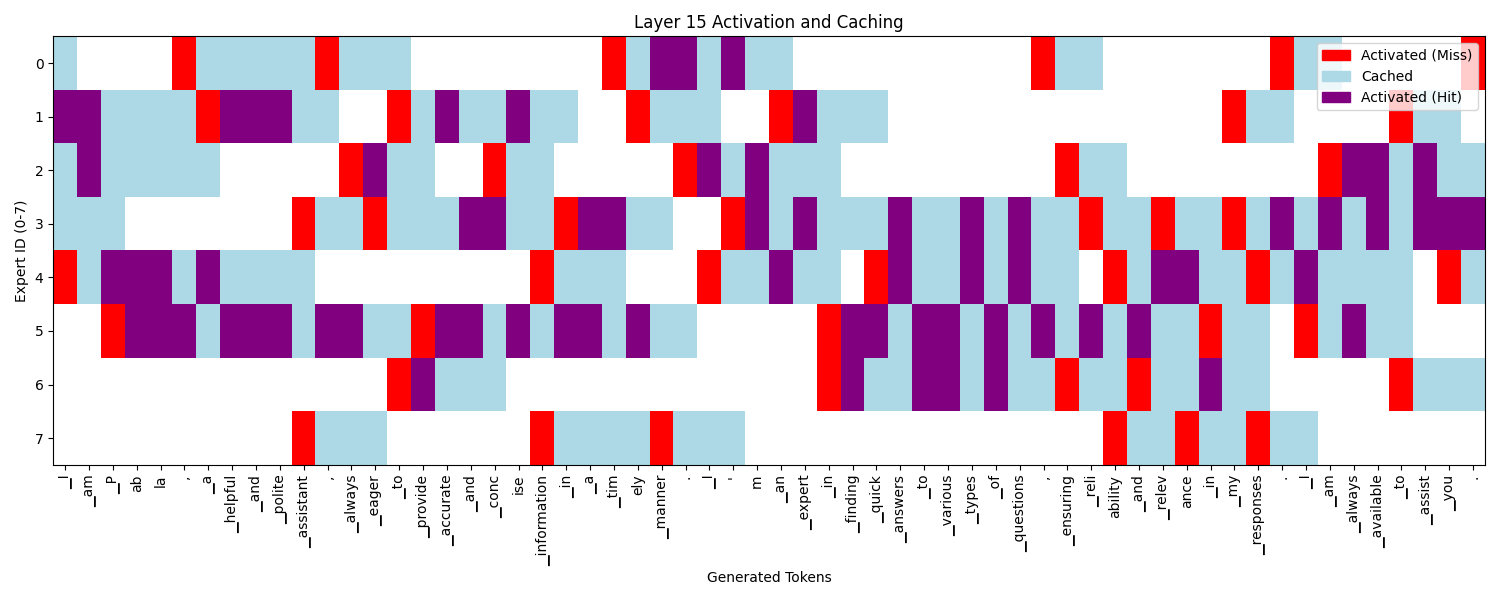}
\caption{The trace of expert activation and LRU cache with cache size=4 for the 16th layer. }
\centering
\label{p1l15}
\end{figure}

\begin{figure}[h]
\includegraphics[width=8cm]{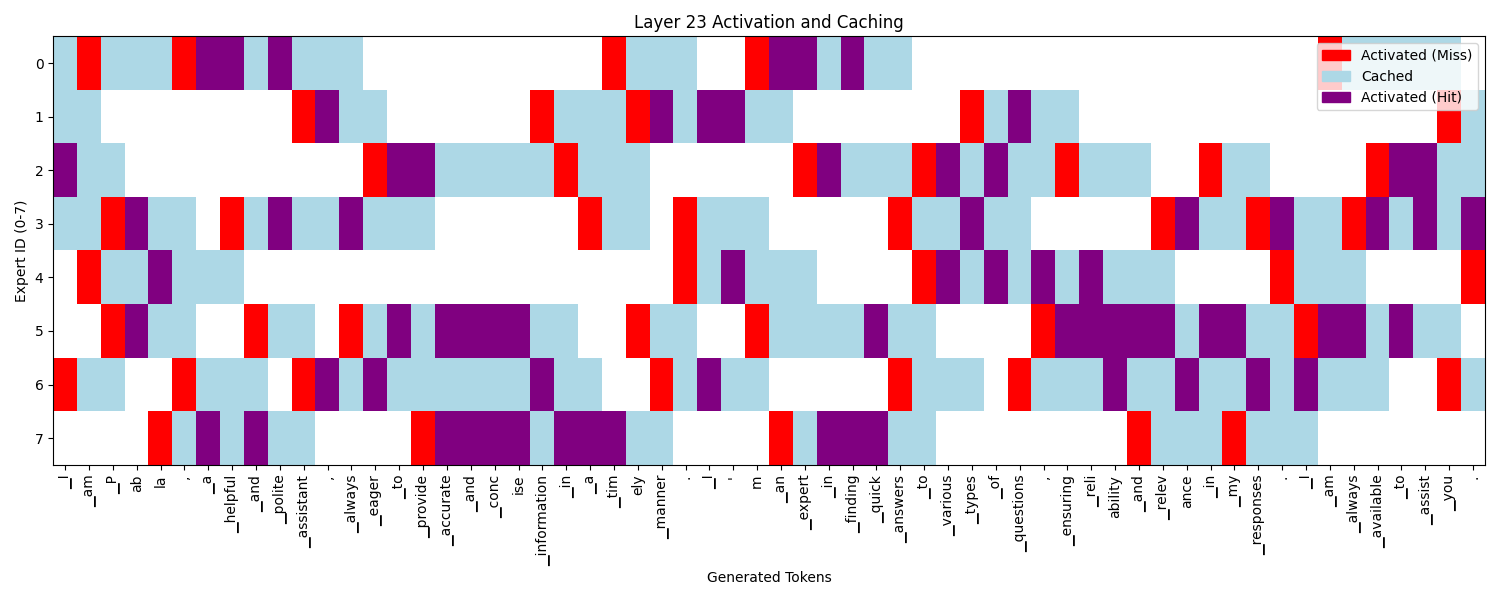}
\caption{The trace of expert activation and LRU cache with cache size=4 for the 24th layer. }
\centering
\label{p1l23}
\end{figure}

\begin{figure}[h]
\includegraphics[width=8cm]{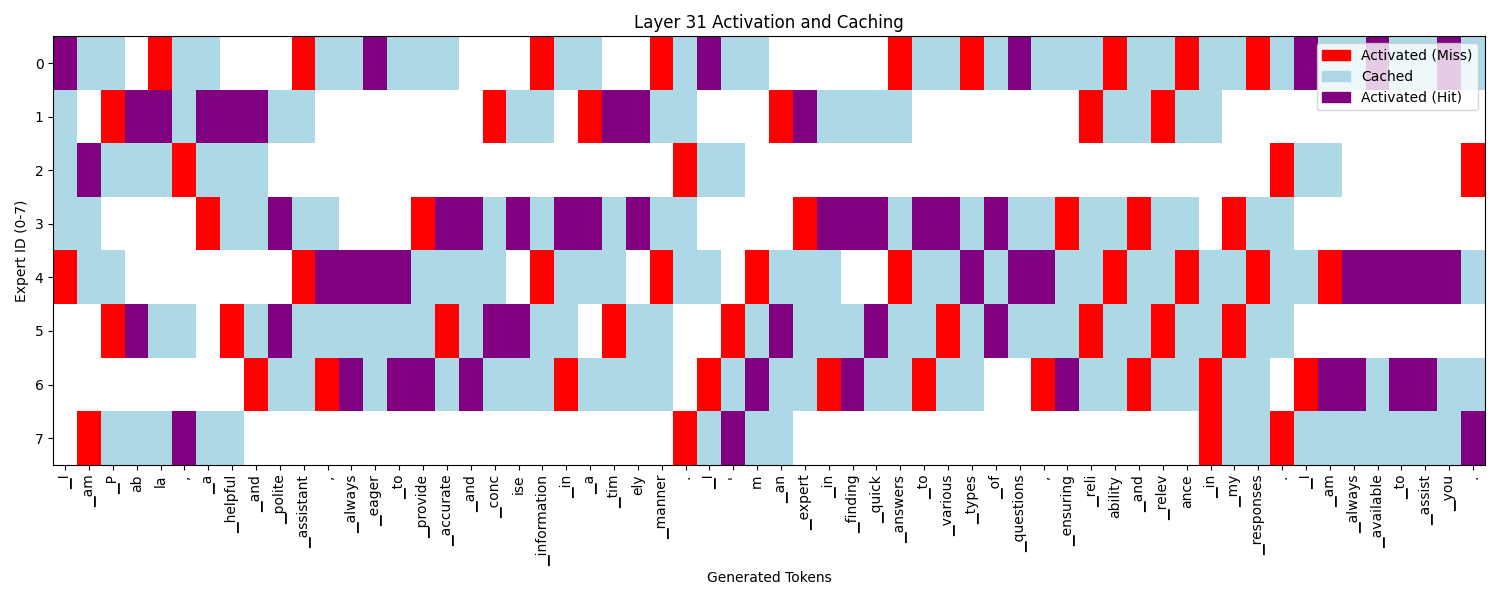}
\caption{The trace of expert activation and LRU cache with cache size=4 for the 32nd layer. }
\centering
\label{p1l31}
\end{figure}

Each figure is the history for one layer, demonstrating how the temporal locality is actually exhibited in reality and how the LRU caching based on this property actually works: Fig \ref{p1l0}, Fig \ref{p1l7}, Fig \ref{p1l15}, Fig \ref{p1l23}, Fig \ref{p1l31}. Note that the tokens in the figures are from the response only, because for a decoder-only architecture, only response generation goes through the attention+MoE forward path.

% When we switch the cache eviction policy to LFU, the speed of token generation remains similar, at about 1.53 tokens per second on a Google Colab T4 GPU. Therefore, we conclude that these changes cannot make cache more effective.

\subsection{LRU Cache Experts Distribution}
\begin{figure}[h!]
    \centering
    \subfloat[\centering 1st Layer]{{\includegraphics[width=3.6cm]{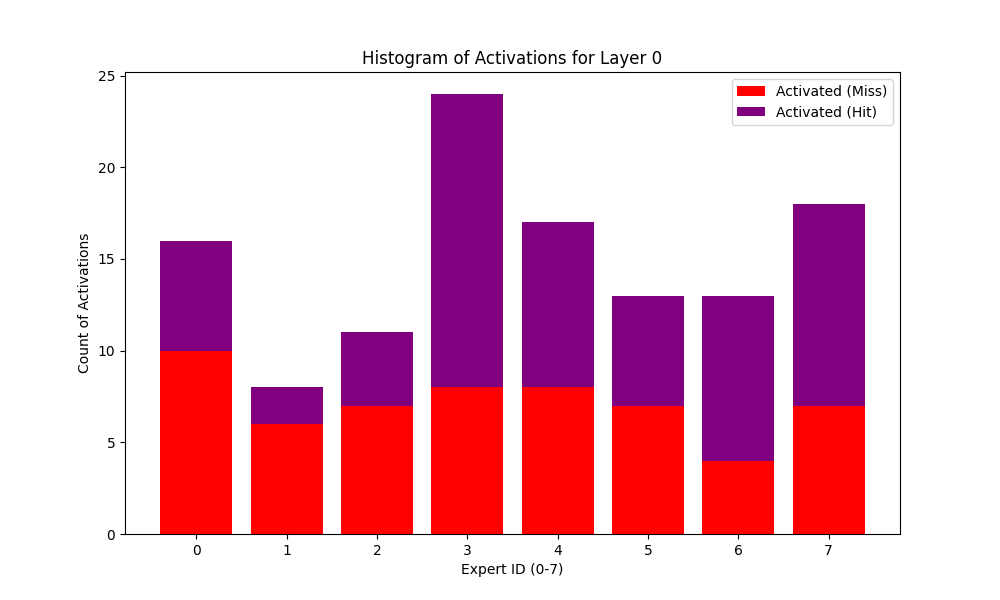} }}
    \qquad
    \subfloat[\centering 2nd Layer]{{\includegraphics[width=3.6cm]{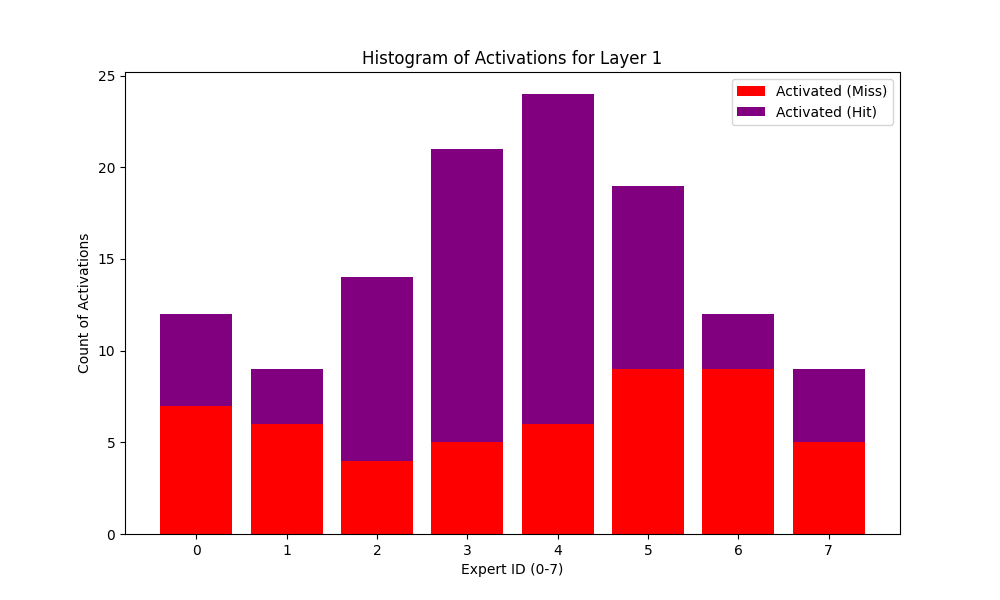} }}
    \qquad
    \subfloat[\centering 7th Layer]{{\includegraphics[width=3.6cm]{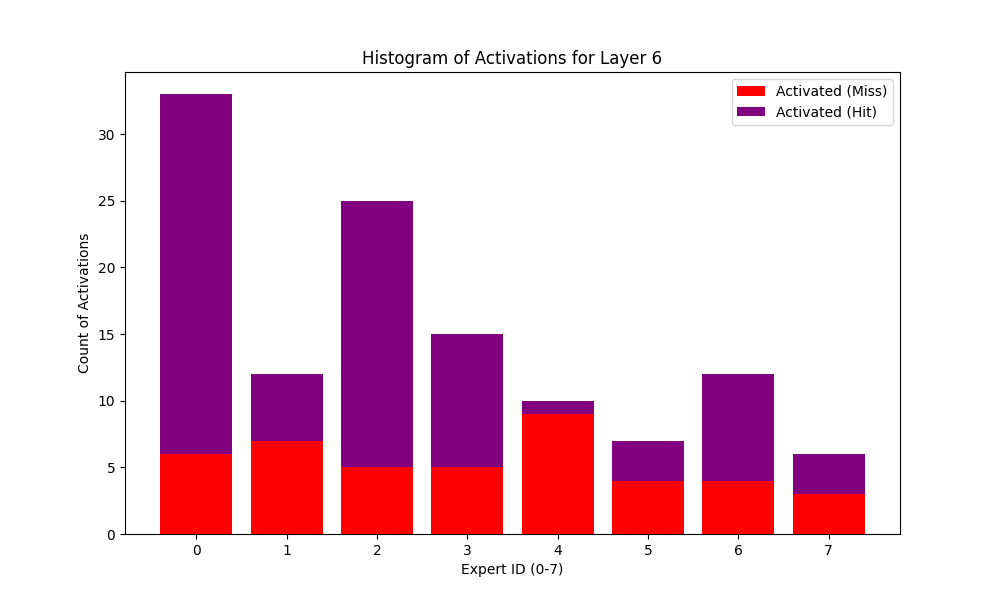} }}
    \qquad
    \subfloat[\centering 8th Layer]{{\includegraphics[width=3.6cm]{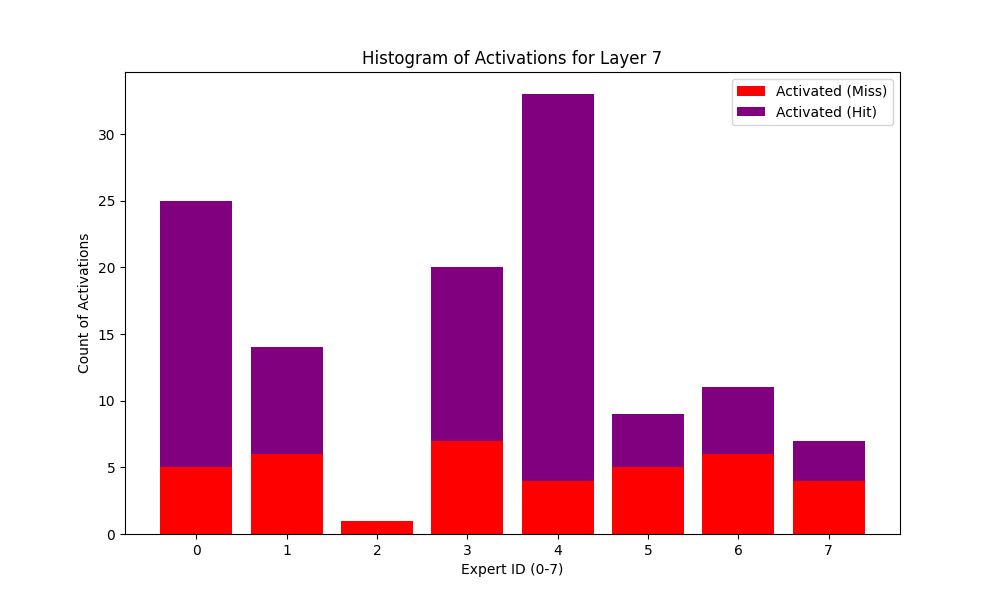} }}
    \qquad
    \subfloat[\centering 15th Layer]{{\includegraphics[width=3.6cm]{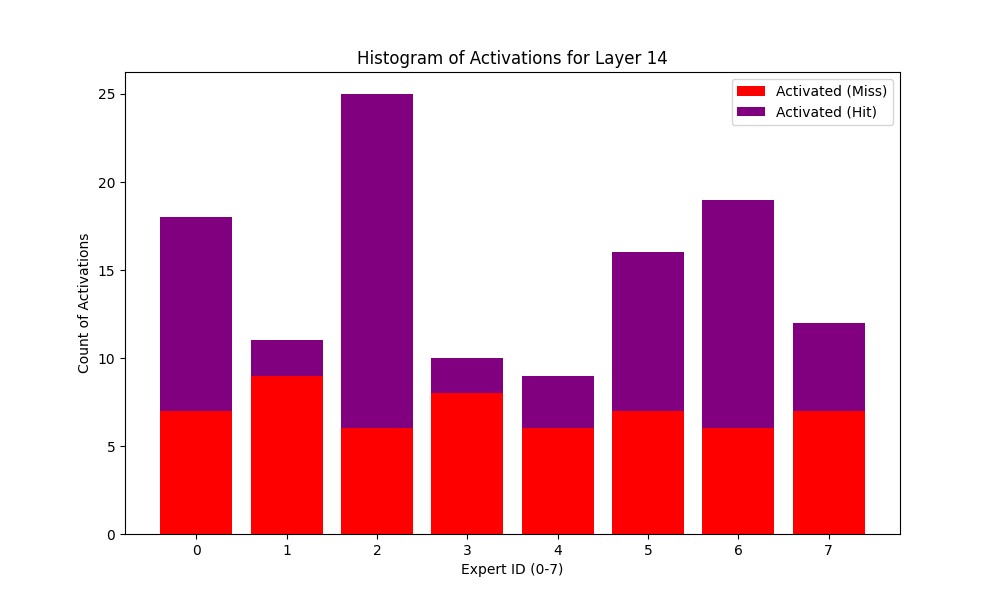} }}
    \qquad
    \subfloat[\centering 16th Layer]{{\includegraphics[width=3.6cm]{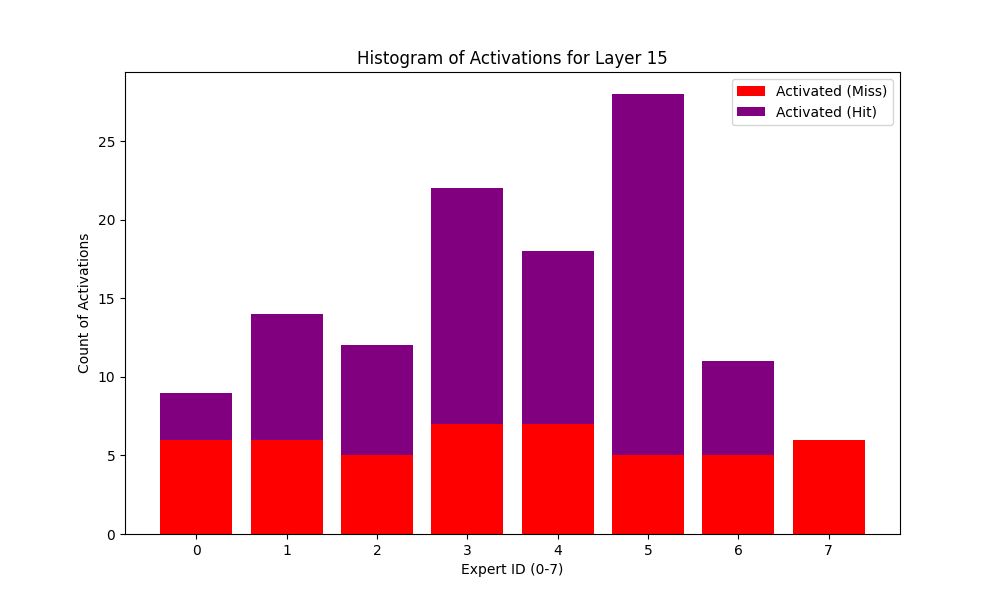} }}
    \qquad
    \subfloat[\centering 23rd Layer]{{\includegraphics[width=3.6cm]{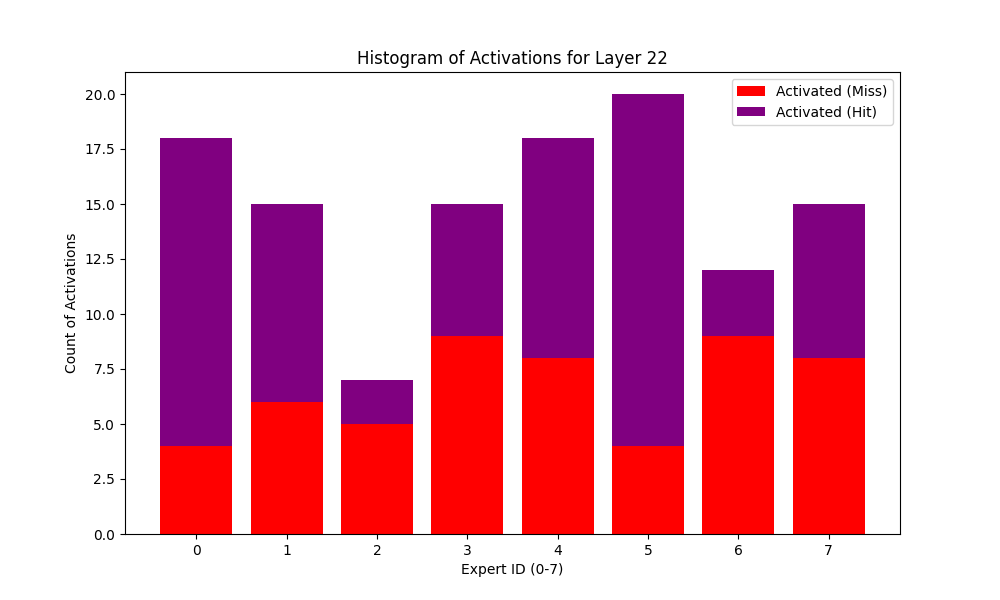} }}
    \qquad
    \subfloat[\centering 24th Layer]{{\includegraphics[width=3.6cm]{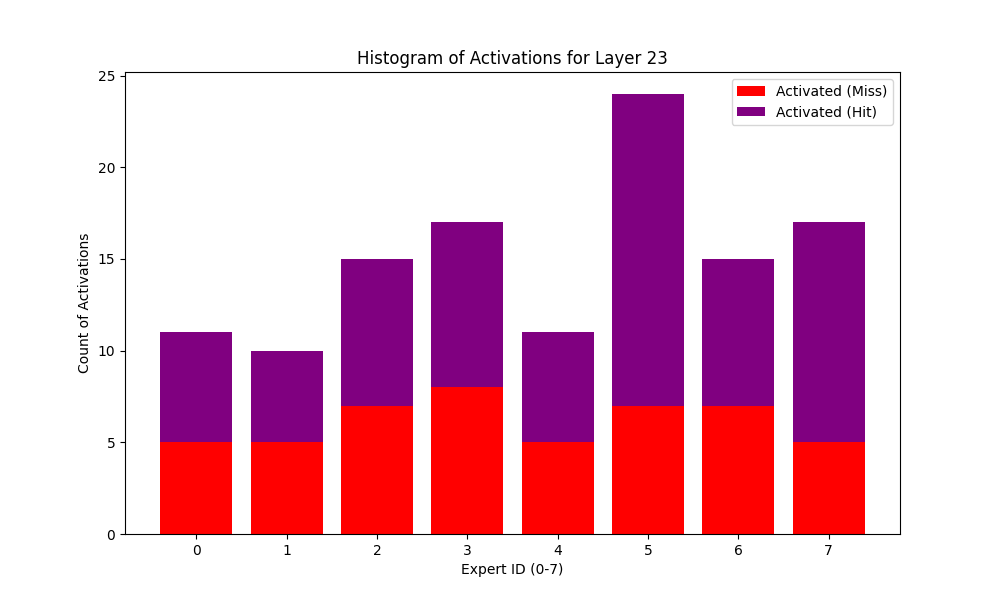} }}
    \qquad
    \subfloat[\centering 31st Layer]{{\includegraphics[width=3.6cm]{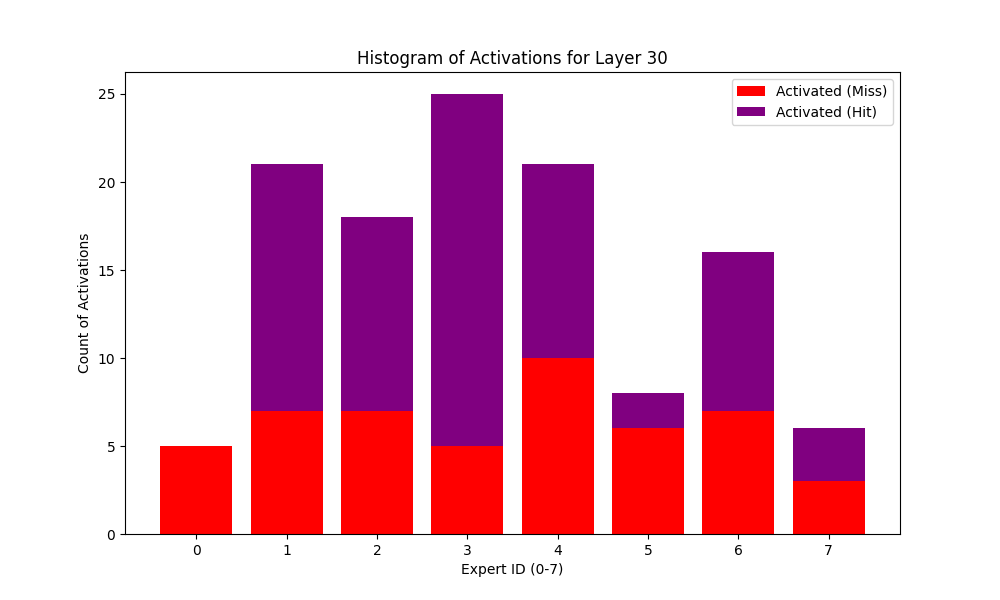} }}
    \qquad
    \subfloat[\centering 32nd Layer]{{\includegraphics[width=3.6cm]{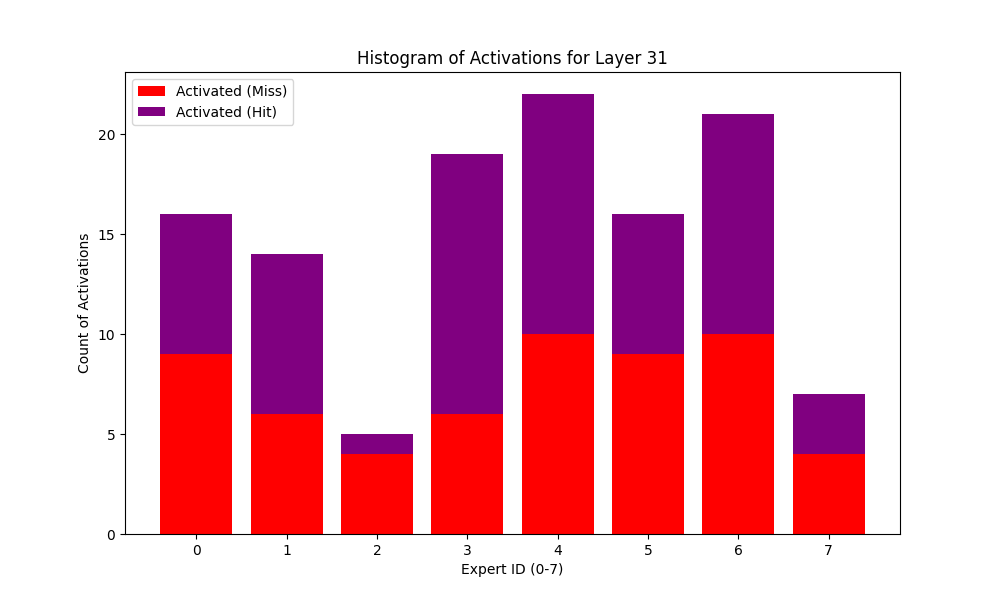} }}
    \caption{Distribution of Activated Experts on the 1st, 2nd, 7th, 8th, 15th, 16th, 23rd, 24th, 31st, 32nd Layer using LRU Caching.}
    \label{fig:hist-lru}
\end{figure}

We analyze the distribution of experts in each layer. There are 32 MoE layers in the Mixtral 8x7B model we experimented on. We select and present the histogram of those layers in the window length of 8 and hop size of 2, specifically 1st, 2nd, 7th, 8th, 15th, 16th, 23rd, 24th, 31st, 32nd layers in Figure \ref{fig:hist-lru}. We have seen that, overall the distributions of activated experts are more skewed in in the middle layers than the layers at the beginning and the end, but hardly have some consistent trends. We observe that these distributions are concentrated in a small number of experts rather than evenly distributed. In some layers, few experts are rarely activated. For example, the number 2 expert on the 8th layer has only been activated once during the entire decoding process. Inspired by this uneven distribution of activated experts, we propose, implement and evaluate the LFU-based expert caching method, as described in Section \ref{lfu-cache}.

\subsection{LFU Cache Performance}
\label{lfu-cache}

We conducted experiments comparing the LRU caching baseline and our proposed LFU method on the same prompt in Section \ref{lru-cache}. Table \ref{tab:speed-comparison} shows the performance comparison, and we observe that the proposed LFU caching method outperforms the LRU baseline on all computation hardware in our experiments. Especially on A6000 GPU, LFU is 84.6\% faster than LRU. This shows the effectiveness of the proposed method, selecting experts to activate by maintaining and based on frequency is better than choosing the recently used experts. We further analyze the expert activation by calculating the precision and recall for the hit activation and missed activation. We see LFU reaches 29.9\% and 59.8\% on precision and recall, outperforming 29.1\% and 58.2\% of LRU by a small margin. This is consistent with the performance in terms of inference speedup.

\begin{table*}[t]
\centering

\resizebox{0.75\textwidth}{!}{%
\begin{tabular}{lcccccc}
\toprule
\multirow{2}{*}{Caching Methods} & \multicolumn{4}{c}{Tokens Per Seconds (s)} & \multirow{2}{*}{Precision (\%)} & \multirow{2}{*}{Recall (\%)}\\ \cline{2-5} 
& A100 & A6000 & L40  & 3090 \\ \midrule
LRU (Baseline)       & 3.33 & 2.34  & 4.17 & 3.07 & 29.1 & 58.2 \\
LFU (Proposed)       & \textbf{3.64} & \textbf{4.32}  & \textbf{4.65} & \textbf{3.09} & \textbf{29.9} & \textbf{59.8} \\ \bottomrule
\end{tabular}%
}
\caption{The Performance Comparison of LRU and LFU on Tokens Per Second Across Multiple Computation Hardwares, and Precision and Recall of Predicted Experts Activation}
\label{tab:speed-comparison}
\end{table*}

\begin{figure}[h]
\includegraphics[width=8cm]{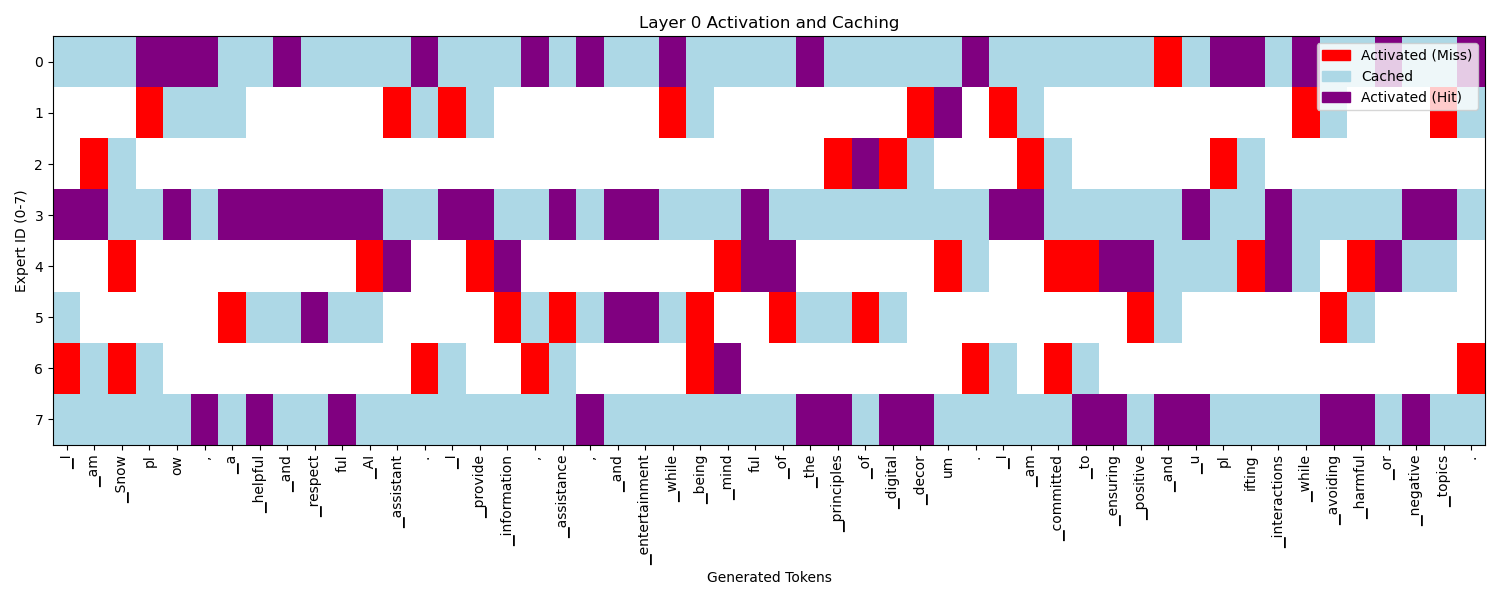}
\caption{The trace of expert activation and LFU cache with cache size=4 for the 1st layer. }
\centering
\label{p1l0-lfu}
\end{figure}

\begin{figure}[h]
\includegraphics[width=8cm]{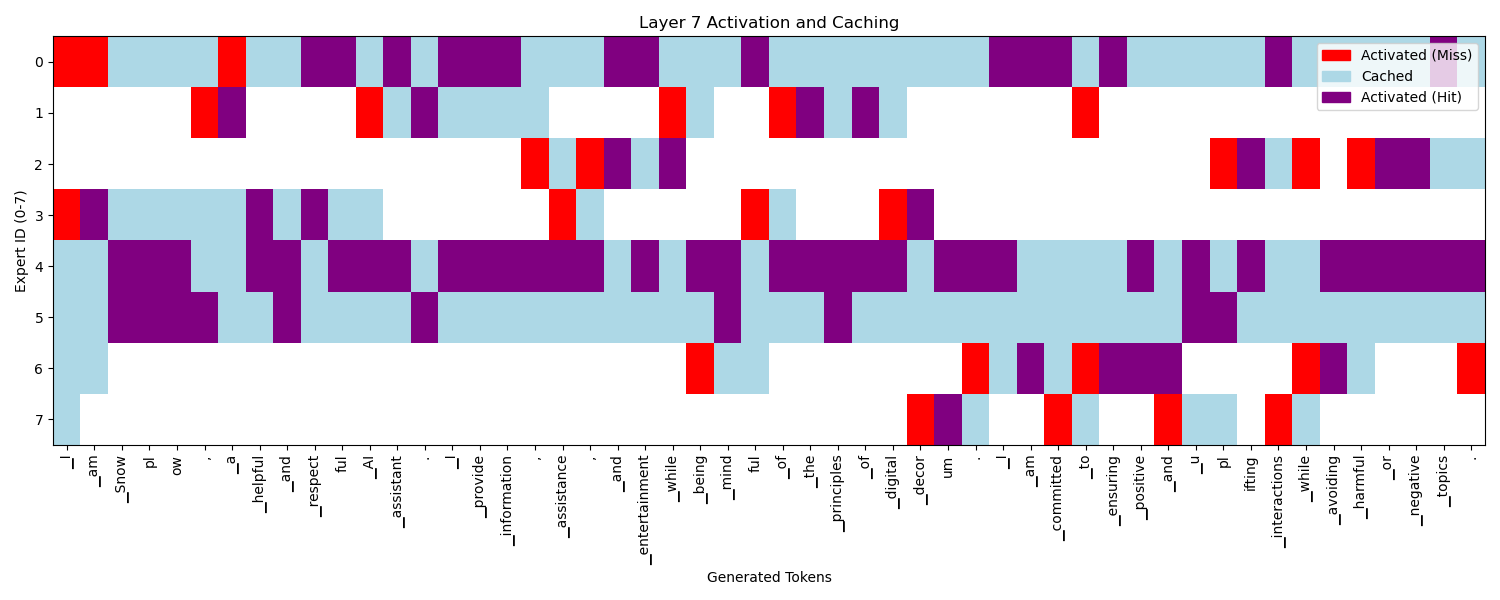}
\caption{The trace of expert activation and LFU cache with cache size=4 for the 8th layer. }
\centering
\label{p1l7-lfu}
\end{figure}

\begin{figure}[h]
\includegraphics[width=8cm]{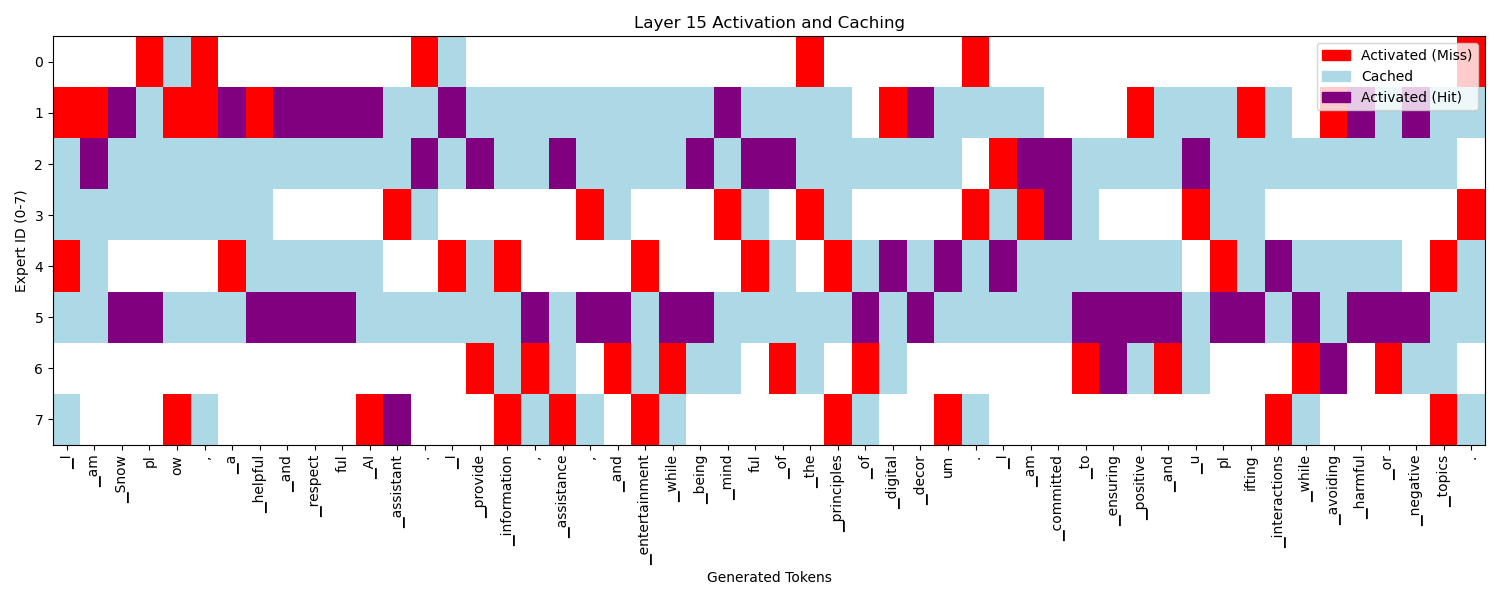}
\caption{The trace of expert activation and LFU cache with cache size=4 for the 16th layer. }
\centering
\label{p1l15-lfu}
\end{figure}

\begin{figure}[h]
\includegraphics[width=8cm]{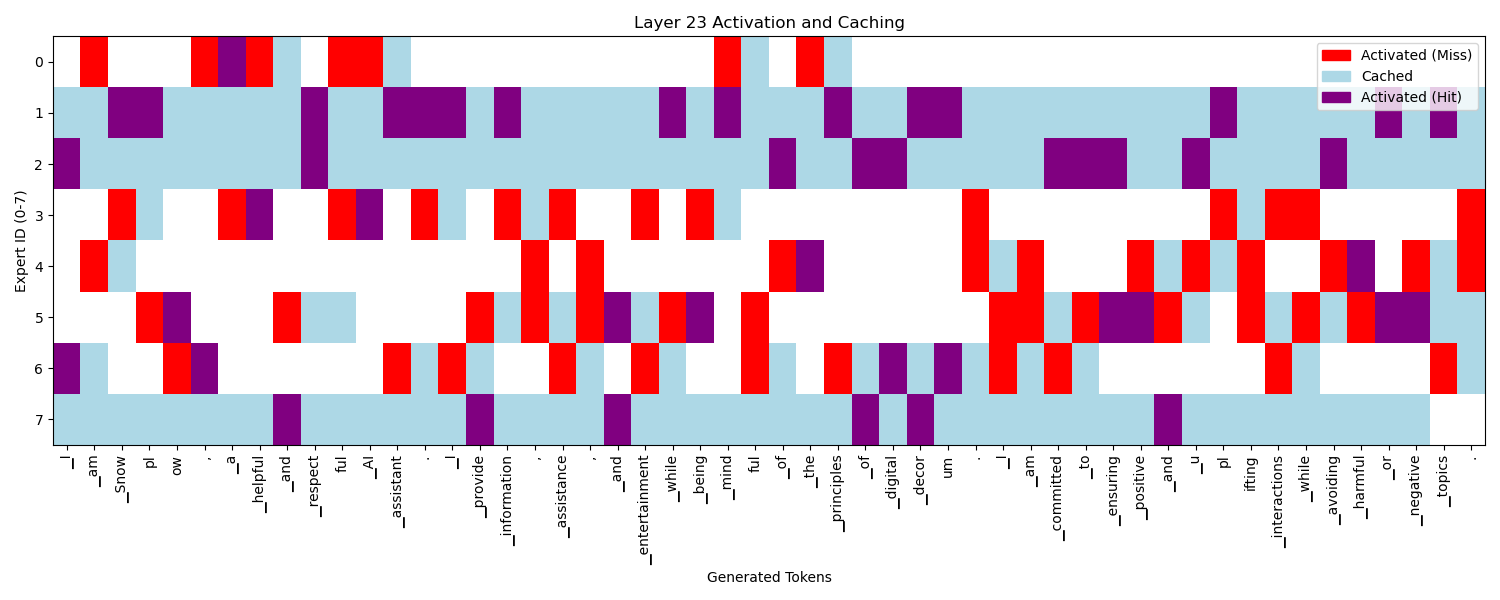}
\caption{The trace of expert activation and LFU cache with cache size=4 for the 24th layer. }
\centering
\label{p1l23-lfu}
\end{figure}

\begin{figure}[h]
\includegraphics[width=8cm]{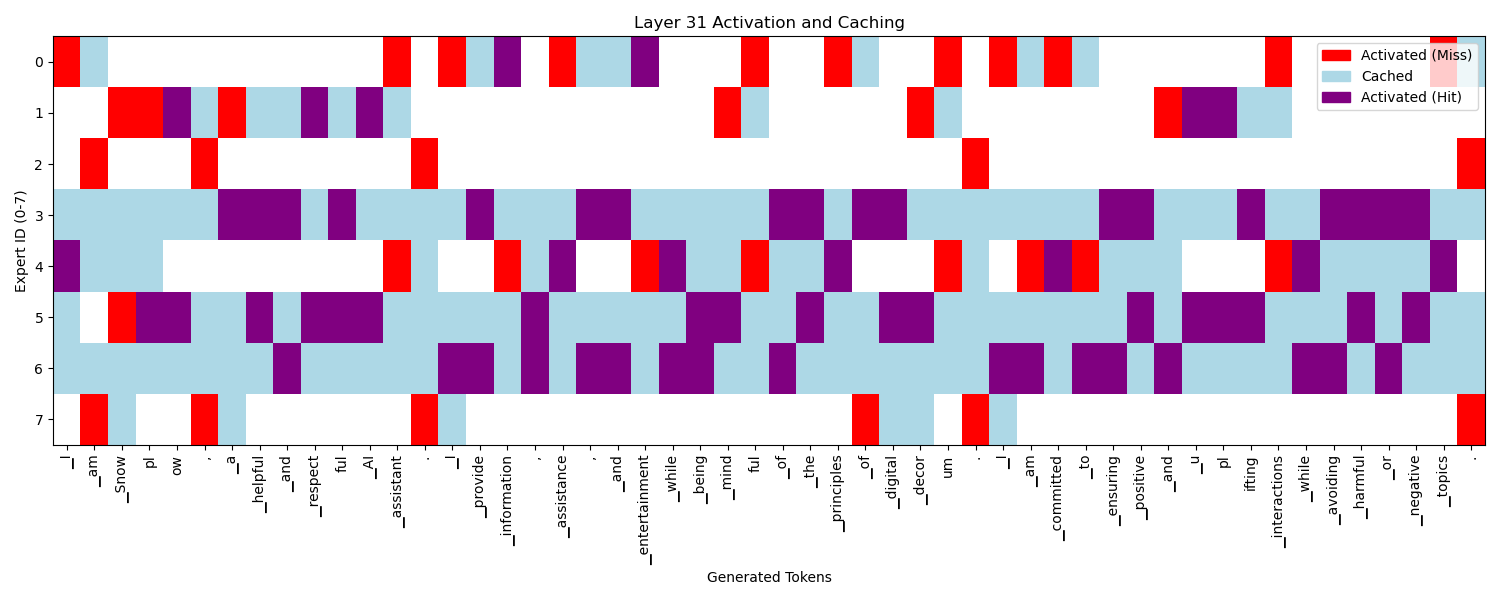}
\caption{The trace of expert activation and LFU cache with cache size=4 for the 32nd layer. }
\centering
\label{p1l31-lfu}
\end{figure}

The activation and caching patterns were visually represented through plots \ref{p1l0-lfu}, \ref{p1l7-lfu}, \ref{p1l15-lfu}, \ref{p1l23-lfu}, and \ref{p1l31-lfu}. It is worth noting that some experts remain in the cache throughout all tokens, showing earlier but more frequent uses of the experts are favored over recent contextual relevance.

\subsection{Speculative Expert Loading Performance}

The precision and recall for speculative expert loading are both 84.6\%(explained in next paragraph) for the prompt ``Introduce yourself, limit your response in 50
words''. This means most of the speculations are correct, so the algorithm can successfully avoid loading an expert in most cases, which indicates huge performance boost.

\begin{figure}[h]
\includegraphics[width=8cm]{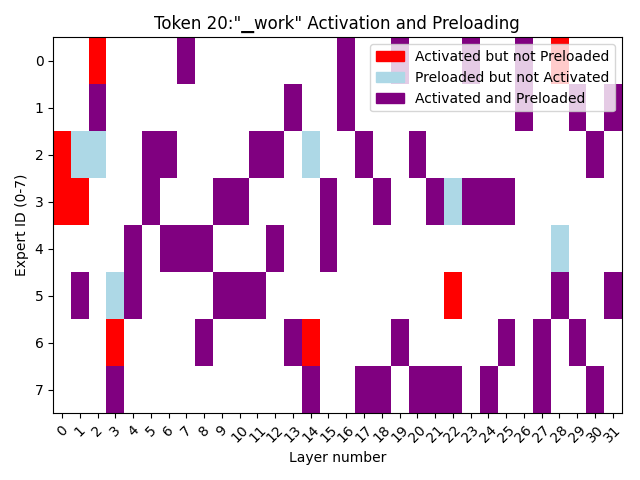}
\caption{The trace of speculative expert loading at all layers for one token.}
\centering
\label{spec_t20}
\end{figure}

\begin{figure}[h]
\includegraphics[width=8cm]{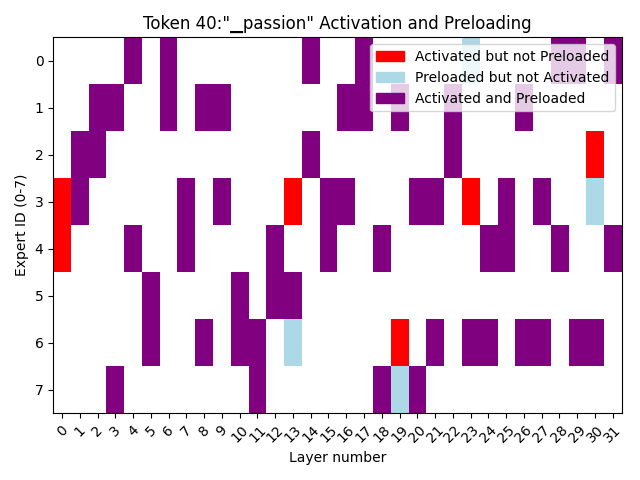}
\caption{The trace of speculative expert loading at all layers for another token.}
\centering
\label{spec_t40}
\end{figure}

To understand why precision is exactly the same as recall: in our calculation, false positive is defined as a guessed expert is not actually activated,  false negative is defined as an activated expert is not guessed, and true positive is defined as an activated expert is guessed. When performing speculative loading, every incorrect guess means it loads an expert that will not be activated, and the expert that should be activated is not guessed: one false positive and one false negative. Therefore, false positive will always be the same as false negative no matter how many of the guesses are incorrect, hence recall is always equal to precision. We present some of the traces for speculative preloading performance(Fig \ref{spec_t20}, Fig \ref{spec_t40}). As shown in the figures: 1. most of the blocks are purple(true positive). 2. number of blue blocks(false positive) is always the same as the number of red blocks(false negative) - excluding the red blocks at the first layer(it's not possible to guess for the first layer, so although the blocks are marked as red, they are not used to compute statistics).

\section{Conclusion}
\subsection{Takeways and Future Directions}
\label{takeways}
\begin{itemize}
    \item \textbf{Caching Algorithm}:
    We find that LFU yields a much faster generation speed than LRU, but its precision and recall are only a little higher than LRU. In addition, both caching algorithms are far from perfect. Our traces demonstrate that both LRU and LFU have a lot to improve. What we learn from LFU is that we cannot allow an expert to be unevictable just because it is popular. Some combination of popularity and unused count might be a better option. 
    \item \textbf{Speculative Expert Preloading}:
    Speculative expert preloading has been demonstrated to have huge potential to speed up inference. However, deploying speculative preloading will cause total amount of parameters transferred to increase as long as there is an incorrect guess, because at the next layer the wrong expert must be swapped with the correct one in main memory. It also competes for the bandwidth with the current layer's expert loading. Overlapping the preloading with computation might resolve these problems, but overlapping itself is a complex topic that we do not dive into for this work. In addition, speculative preloading also occupies the cache space of the next layer, hence it might impact caching performance, however given that precision and recall for speculative loading are much higher than caching, it might be worth to yield space for speculative loading. We believe this is an exciting direction to work in.
    \item \textbf{Overlapping}:
    Based on caching traces, we considered some other optimizations such as learning-based prediction trained from a large dataset of activation history, or early eviction on experts that have not been used for a long time period. However, overlapping is not only important for speculative expert preloading, many other preloading/prefetching techniques like the ones mentioned above, can produce negative rather than positive effects because they add extra transfer beyond necessary swap upon cache eviction. It is necessary to consider overlapping transfer with computation if future researchers intend to implement such methods.
    \item \textbf{Expert Activation}: Although the temporal locality does exist, it is not strong. Compared to temporal locality, expert imbalance is much stronger. From the traces, we can observe that within a sequence, popular experts are selected not always continuously, but still frequently, in intervals. This means that expert imbalance is not necessarily due to proximity between tokens, but also semantic similarity between tokens within a sequence even if they are far away from each other. The context at a larger scale might be a more influential factor than the sequence at a smaller scale for expert selection. This is worth further exploring and may inspire new architecture, for example using only a few popular experts for all tokens in a certain length of sequence might not hurt performance much - a pruning method.
    \item \textbf{MoE Architecture}: The high accuracy of speculative preloading raises a question: Are all layers important? If the activation is the same no matter if it's based on this layer's hidden states or the previous layer's hidden states, it probably means that discarding some layers would not hurt model quality much. It is worthwhile to explore new architecture from this observation, . 
\end{itemize}

\subsection{Limitations}
\begin{itemize}
    
    \item \textbf{Model Specific}:
    Currently, we only work on Mixtral 8x7B. If the model is changed, depending on the model size and configuration, this method might require more compute resources than free-tire Colab.

    \item \textbf{Fixed Dataset Testing}:

    One of the limitations of our experiment is the use of a fixed dataset to assess the performance of the updated LFU caching mechanism in a Mixture of Experts (MoE) system. While testing on a larger dataset of prompts allows us to compute the average performance metrics reliably, it may not fully capture the variability and real-world unpredictability of access patterns. Expert models might exhibit different behaviors under varied workload conditions which are not represented in a fixed dataset.

    To mitigate this limitation, future experiments could incorporate dynamic datasets that evolve over time or datasets that are specifically designed to simulate different types of real-world usage patterns. Additionally, it would be beneficial to test the system under a range of conditions, including varying cache sizes and different types of expert models to understand how LFU caching performs under different scenarios.

    \item \textbf{Consistency in Generated Outputs}

    Another limitation arises from the variability in the outputs of the MoE system during experimental trials, especially when comparing performance across different caching strategies. In our initial setup, each run may produce different outputs due to the stochastic nature of the models involved, particularly if the temperature parameter during inference is not controlled. This variation can affect the comparability of performance metrics such as inference speed and caching precision/recall.

\end{itemize}

\nocite{langley00}
\bibliography{paper}
\bibliographystyle{mlsys2024}

%%%%%%%%%%%%%%%%%%%%%%%%%%%%%%%%%%%%%%%%%%%%%%%%%%%%%%%%%%%%%%%%%%%%%%%%%%%%%%%
%%%%%%%%%%%%%%%%%%%%%%%%%%%%%%%%%%%%%%%%%%%%%%%%%%%%%%%%%%%%%%%%%%%%%%%%%%%%%%%
% SUPPLEMENTAL CONTENT AS APPENDIX AFTER REFERENCES
%%%%%%%%%%%%%%%%%%%%%%%%%%%%%%%%%%%%%%%%%%%%%%%%%%%%%%%%%%%%%%%%%%%%%%%%%%%%%%%
%%%%%%%%%%%%%%%%%%%%%%%%%%%%%%%%%%%%%%%%%%%%%%%%%%%%%%%%%%%%%%%%%%%%%%%%%%%%%%%
\appendix

\end{document}